# A combination of 'pooling' with a prediction model can reduce by 73% the number of COVID-19 (Corona-virus) tests

Tomer Cohen, Lior Finkelman, Gal Grimberg, Gadi Shenhar, Ofer Strichman, Yonatan Strichman, Stav Yeger

Technion, Israel Institute of Science.

Contact: ofers@technion.ac.il

1. Background

The search for carriers of COVID-19 is done primarily through testing using RT-PCR's. These tests are the most common way to empirically identify carriers of the virus, and urgently need to be conducted on a large scale. Today, patients are granted a test if deemed necessary by the government and are carried out individually, i.e., every sample is tested separately.

The problem is that the number of samples gathered today supersedes the amount of tests that can be conducted daily; Moreover, the world-wide shortage in equipment and resources prevents a much-needed increase in the number of daily tests. As a result, the testing system today is at full capacity, and falls short of the need.

Two recent developments are relevant to the solution that we describe here:

1. **Data regarding tests** and the patients behind them has been gathered (over 120,000 tests in Israel as of Mid. April, 2020) that enables us to build a prediction model of who is likely to be positive;
2. A new study [1] has shown that it is possible to combine up to 32 samples in one 'pool' and identify whether **at least one of them** is positive with a single test. Pooling in general (in the context of other tests) is an old idea due to Dorfman [3].

A simple calculation shows that creating random pools of samples will not be efficient in identifying positive patients, with the current rate of positive samples in Israel -- ~8%. With this rate, less than 7% of the pools will succeed, and the rest will have to be re-tested one by one. Our method solves this problem by identifying those samples that have a much smaller chance of being positive and putting them in one pool. Furthermore, it recommends the optimal size of the pool, given the probability that samples in that pool are positive.

2. The suggested method

By using the meta-data of the tests, which was gathered by the ministry of health, we created an algorithm (based on machine learning, and specifically a 'neural network'), that predicts the probability of a patient being negative or positive for the virus based on his/her data: whether they cough, has a sore throat, shortness of breath, or a headache, as well as his/her age group, gender, and likely cause for the sickness. Based on this data our model predicts **with an accuracy of 95.5%** the outcome[1].

For each patient, the outcome of the neural network is a number indicating the likelihood that this patient is positive. Hence by sorting the patients according to this value, we can pool together tests with almost a uniform probability of being positive.

In appendices A – C, we survey three different ways to use this data – the first-of-which is a known method due to Dorfman[3] - which we'll call here `single pooling' (Appendix A): according to this method, if the pool turns out to be positive, all samples in that pool need to be re-tested individually, as shown in the following diagram:

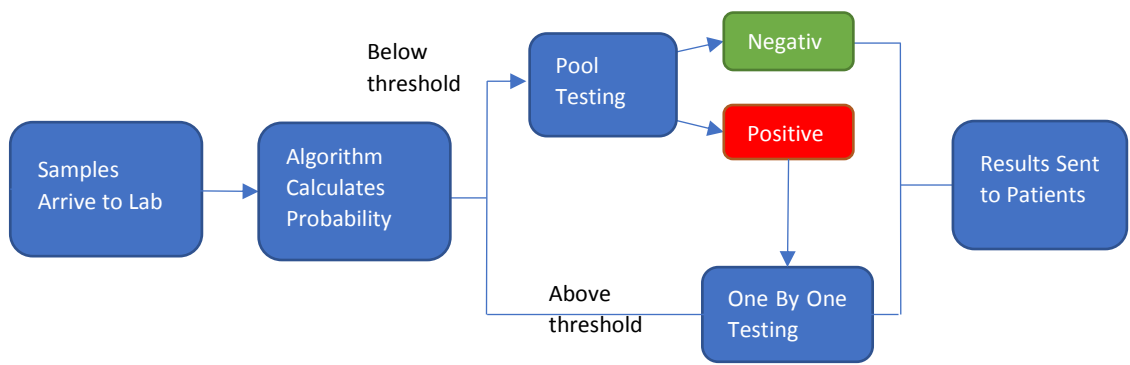

The expected number of tests can be calculated, based on the probability $p$ of the samples in the pool to be positive, and the size $n$ of the pool (see Appendix A for details). We show that for each value of $p$ there is an optimal pool size. For example, for patients with 1% probability of being positive, the optimal pool size is 11. For probabilities higher than some threshold, it is not cost-effective to use pooling at all.

We calculated that if this method had been used on all tests carried thus far in Israel (see Appendix E for data), we would reach full accurate classification of all the patients with about **33% of the tests.** The best method that we found is called '2D-pooling' (or 'matrix pooling') [6][7], and is described in Appendix C – it can do the same thing with only **27% of the tests.** Appendix D includes a table comparing the required number of tests per patient with the various methods, as a function of $p$, and also a comparison to a recently introduced method called double-pooling [4].

3. Challenges: changing the process, and the overhead of re-testing.

Our **solution does not require new equipment**. It requires, however, a change in the current process. First, the samples corresponding to patients that are selected for pooling, will have to be duplicated for potential repetition of the test, in case the pool turns out to be positive.

---

[1] The dataset that we used was published by the MOH. It has a known bias which likely distorts the result: positive cases are much more likely to include the clinical data mentioned above than negative ones, since for negative cases this data was not always entered retroactively. In itself this may have contributed to the success of the prediction model, since the very fact that there is no clinical data is a good predictor of the result. This problem will disappear if the data collection process will improve. The numbers that we present here will likely not be as good in practice because of this reason. A detailed discussion of this matter appears in [5].

Second, the tests that are currently given to the lab in some arbitrary order, will have to be resorted, based on the recommendation of our system.

4. Additional objectives
    The total number of tests is not the only objective. One should also consider
    - The number of test iterations. Each such test takes time (several hours on the PCR machine), which leads to a delay in the response time to the patient.
    - The amount of sample duplication. If a sample will **potentially** need to be retested, it has to be duplicated.

    In Appendix D we consider these objectives when comparing the various methods.

# Appendices

In appendices A – C we describe different pooling methods, with a decreasing number of expected tests. Appendix D compares the efficiency of the various methods. The raw data can be downloaded from [2].

Appendix A: the single pooling method [3]

In this model a sample can only be part of a pool once. That is, if the pooled sample fails, then all samples in the pool are re-tested individually.

Let $P$ be the probability of a test to be positive, $n$ the pool size, and $m$ the population size. The expected number of tests, as a function of the probability of a test to be positive and the pool size is calculated as follows:

$$E(\# \ tests) = \{(1 - (1 - P)^n) \cdot (n + 1) + (1 - P)^n\} \cdot \frac{m}{n} \qquad (1)$$

In the case of a failure there are $n + 1$ tests (left part of (1)) for $n$ samples, and in case of success there is a single test for $n$ samples (right part of (1)). The overall number of pools is $m/n$.

Let us take two numeric examples: for $p = 0.1, n = 12, m = 32$, the average number of tests is 25.62. For $p = 0.01$ and the same values of $n, m$, the expected number of tests drops to 6.4. It is clear that from some threshold for the value of $p$, it is not cost-effective to use pooling at all.

The plots in Figure 1 are based on (1), divided by $m$, so it reflects the expected number of tests per patient.

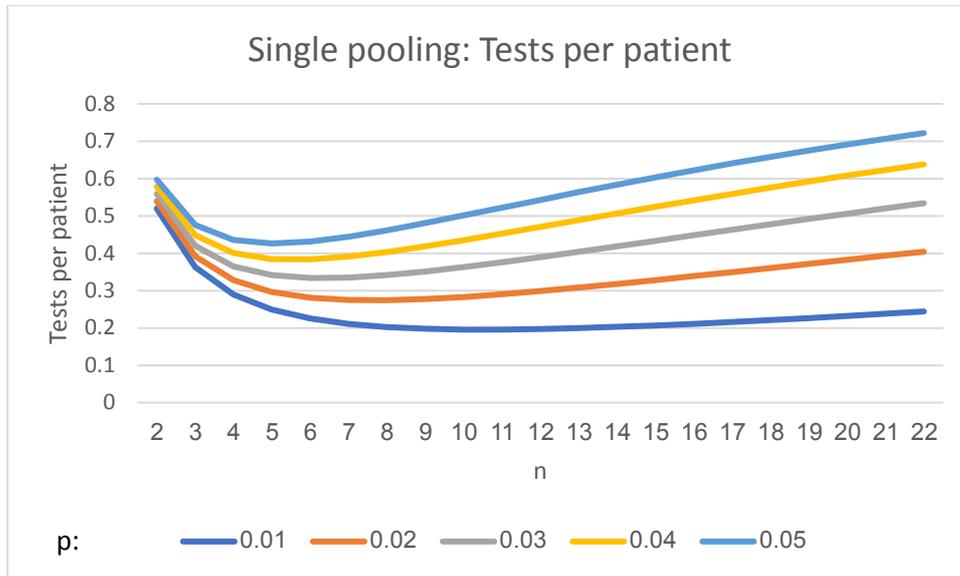

*Figure 1*

It is evident that for each probability, there is an optimal pool size – the size that brings to minimum the expected number of tests. For example, for $p = 0.01$, the optimal pool size is 12, and we saw earlier that this implies 0.19 tests per patient.

Appendix B: The binary-tree method

One can extend the idea presented in Appendix A to multiple levels of pooling. This is sometimes called 'multi-stage pooling' [8]. That is, given a pool of size $n$ that fails, split it to two and retest, until reaching the leaves of the search tree. A small optimization is achieved as follows. Suppose that a node at level $i$ in the binary tree is positive, and we then check, e.g., the left child node at level $i - 1$. If that node is negative, not only that we can skip the whole subtree under that node, we can also skip the other node at level $i - 1$, because it is bound to be positive. The plots in Figure 2 below are based on simulations, including this optimization.

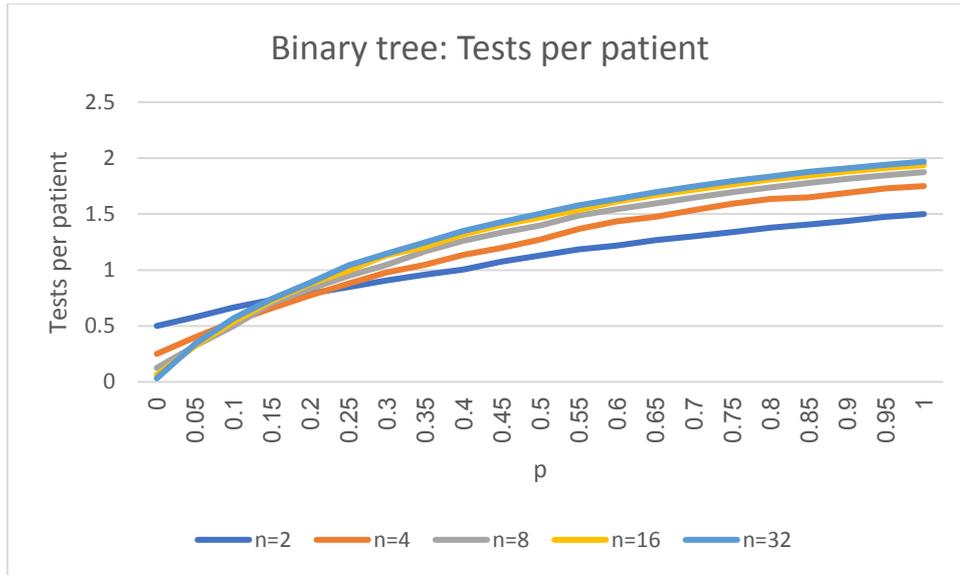

*Figure 2*

Some of the data from our simulation with $m = 32$ appear in Table 1 below, where the optimal pool size (the value of $n$) is highlighted for each value of $p$. One may observe that above a certain probability ($p = 0.35$) pooling is not cost-effective.

*Table 1*

| p -> | 0.0 | 0.05 | 0.1 | 0.15 | 0.2 | 0.25 | 0.3 | 0.35 | 0.4 |
|---|---|---|---|---|---|---|---|---|---|
| n=1 | 32.0 | 32.0 | 32.0 | 32.0 | 32.0 | 32.0 | 32.0 | 32.0 | 32.0 |
| n=2 | 16.0 | 18.5 | 21.3 | 23.5 | 25.3 | 27.1 | 29.1 | 30.6 | 32.2 |
| n=4 | 8.0 | 12.7 | 17.1 | 21.1 | 24.8 | 28.2 | 31.4 | 33.5 | 36.3 |
| n=8 | 4.0 | 10.4 | 15.9 | 22.5 | 26.6 | 30.3 | 33.5 | 37.4 | 40.4 |
| n=16 | 2.0 | 10.3 | 17.2 | 23.2 | 28.0 | 31.9 | 36.2 | 38.6 | 42.2 |
| n=32 | 1.0 | 10.8 | 18.2 | 23.8 | 28.4 | 33.3 | 36.7 | 40.0 | 43.2 |

Comparing binary search to 'single-pooling' (Appendix A), one can observe that it leads to a smaller number of expected tests. For example, for $p = 0.05$ with pool size of 16, the expected number of tests (for $m = 32$) is 10.3. On the other hand, with the same value of $p$ single-pooling suggests a pool size of 5, and the corresponding expected number of tests is 13.6.

The main disadvantage of the binary tree method is that it imposes $\log n$ iterations (the height of the binary tree). Furthermore, if the optimization mention above is activated, it doubles the number of iterations (because one cannot test the two sibling nodes simultaneously). See Appendix D for further discussion.

Appendix C: The 2D-pooling method [6][7]

Another method of pooling is based on a matrix of pools. We arrange an $n \times n$ matrix of samples. We then pool together each column and each row, hence $2n$ pools of $n$ samples each. This implies that each sample participates in two pools (i.e., row and column pools). Suppose that $k$ out of the $n^2$ samples are positive. This means that in the worst case $2k$ pools will test positive (as each such sample makes a separate row and column positive), and correspondingly there will be $k^2$ samples that are potentially positive and hence need to be re-tested. There is a limit, however, on the number of additional tests: it cannot surpass $n^2$. Hence the worst case can only happen $n$ times.

**Example 1**   The grid (also can be seen as a matrix) in Figure 3 is for $n = 5$. Each of the 25 junctions is a sample. Correspondingly, 10 pools will be tested. For example, the 5 samples in the first row form a pool, and the 5 samples in the first column form another pool. Suppose there are 2 positive cases from those 25 samples, and they are located at (2,2) and (4,4) (marked in red). Correspondingly, row-pools 2, 4, and column pools 2,4 are positive (marked by 'X'). But this leaves us with 4 candidate samples, which must be re-tested individually. So altogether, in this case we need 14 tests, for 25 people. A more fortunate case is when the positive cases happen to be on the same row or column. For example, if those two positive cases were on the same row, we would not need more tests at all as they would be uniquely identified.

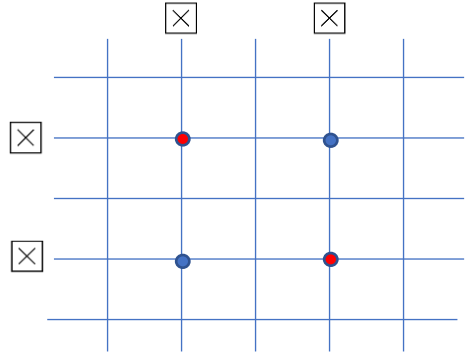

*Figure 3: a grid for Example 1, where the two red dots denote positive cases. These 4 cases have to be retested individually.*

Eq. (2) below expresses the expected number of tests with the 2D-pooling method, in the worst case. As before, $p$ is the probability for a positive test, $m$ is the size of the population, and $n$, in this case, is the number of columns/rows in the matrix. Recall that the worst case is when each positive sample is a singleton on its row and column. In that case the number of added tests is $k^2 = (pn^2)^2$. Hence, we have

$$E(\# \, tests) = (2n + (pn^2)^2)m/n^2 \qquad (2)$$

This expression is relevant only when $k \leq n$. Since $k = pn^2$, this happens when $pn^2 \leq n$, or $p \leq 1/n$. Furthermore, it is relevant only as long as $E(\# \, tests) < m$, which implies that

$$p < \sqrt{(n-2)/n^3}. \qquad (3)$$

Eq. (3) is a strictly tighter bound than the former $p \leq 1/n$.

Plotting (2) for different values of $p$ and $n$ where $m = 400$ (see Figure 4), shows that here, too, for each value of $p$, there is an optimal matrix size.

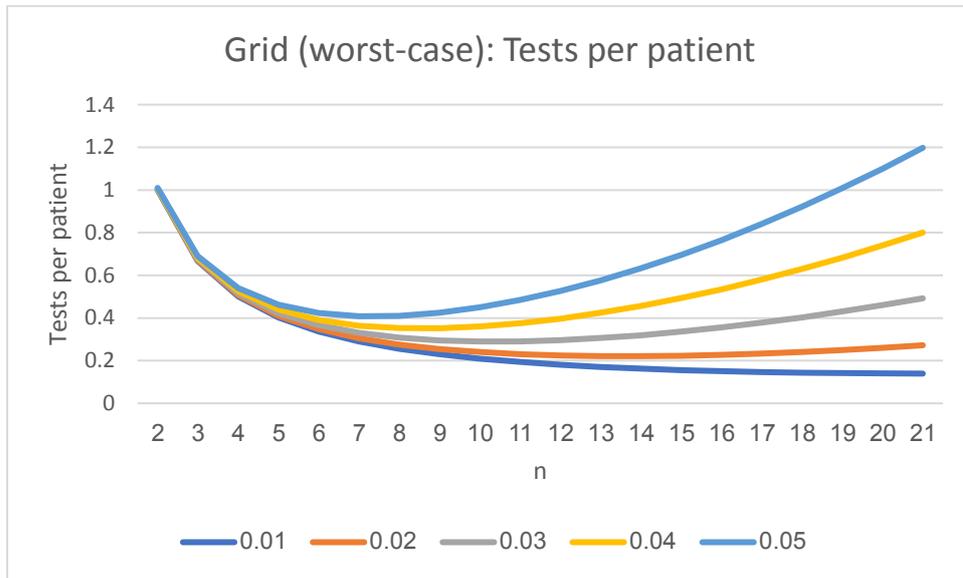

*Figure 4*

We note that for these results to apply, we must have $n^2 \leq m$ (otherwise we do not have enough samples to fill the matrix). For example, for $p = 0.01$ the optimal value of $n$ is 20, which means that we can use the result only if $m \geq 400$ (here the results were computed, as mentioned, with $m =400$).

We also simulated this method, in order to get the actual statistics rather than the worst-case scenario. We included the following (small) optimization in the simulation: suppose that the number of positive rows is $r$. For each positive column, if the first $r - 1$ samples turns out to be negative, there is no need to test the last one, as it is bound to be positive. The same argument applies to each row. In the best-case scenario all the positive cases are in the last row and last column – in such a case we save $2n - 1$ tests (-1 because of the bottom corner). The downside of this optimization is that it adds an iteration. See Appendix D for a discussion.

There are other possible optimizations, which are discussed in [7][8] and we did not include in our simulation. For example: test rows only; if none of them is positive – stop; if exactly one row is positive, check all samples in that row individually; otherwise – continue as usual.

Appendix D: Comparing the methods

Table 2 below compares the three methods in terms of their **tests per patient (TPP)**: the ratio between the expected number of tests and the population size $m$. Note that each cell was calculated with the optimal pool size $n$ for the given value of $p$. For example, for $p = 0.01$ the optimal value of $n$ in the binary-tree method is 11, whereas the optimal value of $n$ in the 2D-pooling method for this probability is 20.

The last line of the table refers to results of simulating the 2D-pooling method, which is expected to be better (and indeed our results show that it is) than the worst-case as formulated in (2).

It is clear from the table that the actual (i.e.., simulated) 2D-pooling method is the most efficient one in terms of minimizing the number of tests.

| p -> | 0.01 | 0.02 | 0.03 | 0.04 | 0.05 |
|---|---|---|---|---|---|
| single-pooling | 0.20 | 0.27 | 0.33 | 0.38 | 0.43 |
| binary tree | 0.10 | 0.17 | 0.23 | 0.27 | 0.33 |
| 2D-pooling (worst-case, according to (2)) | 0.14 | 0.22 | 0.29 | 0.35 | 0.41 |
| 2D-pooling (avg-case, according to simulation) | 0.10 | 0.13 | 0.20 | 0.27 | 0.32 |

*Table 2: The ratio between the expected number of tests and the population size $m$. The smaller the number, the more efficient the method is. These values are calculated with the best value of $n$ for the given probability.*

Finally, let us compare the 2D-pooling method to a recently introduced method called double-pooling [4]. Citing [4]: "given a probability $p$ of a positive test, pick an optimal size $s2(p)$ for the pool size. Divide the population to be tested into non-overlapping pools of size s2 (the division is assumed to be random) twice. Thus, now every patient belongs to two pools and is tested in two parallel rounds, A and B. For every patient if both the pools test positive then test the patient individually. Otherwise consider that patient cleared.". Using their analysis (which we independently verified via simulation), the tests-per-patient of double-pooling is the following:

| p -> | 0.01 | 0.02 | 0.03 | 0.04 | 0.05 |
|---|---|---|---|---|---|
| Double pooling | 0.13 | 0.21 | 0.27 | 0.32 | 0.37 |

Hence both the 2D-pooling method and the binary-tree method are more efficient than double-pooling. Both the 2D-pooling- and the double-pooling method require two steps.

So far we only compared the methods by the expected number of tests per patient. Let us now compare them by the number of iterations, and the number of sample duplication that is necessary (this represents the number of times a patient will be retested in the worst case):

| | Iterations | Sample duplication |
|---|---|---|
| **Single pooling** | 2 | $m$ |
| **binary tree** | $\log n$, (or $2 \log n$ with the optimization). | $\log m$ |
| **2D-pooling** | 2 (or 3, with the optimization). | $m$ |
| **Double pooling** | 2 | $m$ |

Appendix E: the current distribution of samples in Israel

In Figure 5, the horizontal axis shows the value predicted by our neural network, multiplied by 100. The vertical axis shows the accumulated percentage of the population that falls under this value. For example, over 80% of the population are classified by our neural network as having less than 2% chance of being positive. The raw data is accessible from [2].

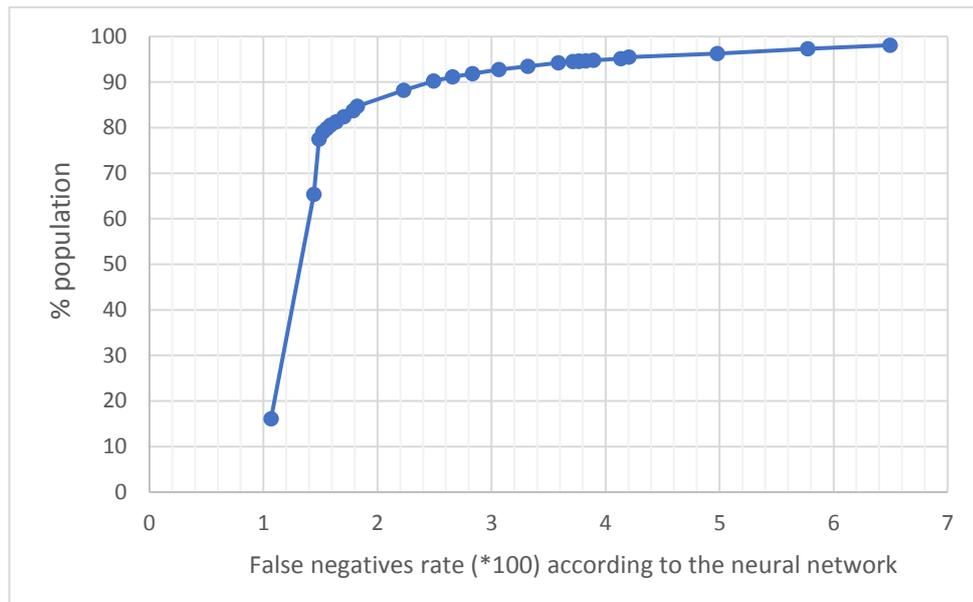

*Figure 5*